\def\year{2020}\relax
\documentclass[letterpaper]{article} % DO NOT CHANGE THIS
\usepackage{aaai20}  % DO NOT CHANGE THIS
\usepackage{times}  % DO NOT CHANGE THIS
\usepackage{helvet} % DO NOT CHANGE THIS
\usepackage{courier}  % DO NOT CHANGE THIS
\usepackage[hyphens]{url}  % DO NOT CHANGE THIS
\usepackage{graphicx} % DO NOT CHANGE THIS

\newcommand{\term}[1]{\textsl{#1}}
\usepackage{makecell,multirow}
\usepackage{amsthm,amsfonts}
\usepackage[ruled,linesnumbered,titlenumbered]{algorithm2e}
\usepackage{graphicx}  % DO NOT CHANGE THIS
\title{Do Not Have Enough Data? Deep Learning to the Rescue! }
\author{Ateret Anaby-Tavor,\textsuperscript{\rm 1}
Boaz Carmeli,\textsuperscript{\rm 1,\thanks{Equally main contributors }}
Esther Goldbraich,\textsuperscript{\rm 1}
Amir Kantor,\textsuperscript{\rm 1}
George Kour,\textsuperscript{\rm 1,2,$^{*}$}
\\ \bf \Large
Segev Shlomov,\textsuperscript{\rm 1,3,$^{*}$}
Naama Tepper,\textsuperscript{\rm 1}
Naama Zwerdling\textsuperscript{\rm 1}
\\
\textsuperscript{\rm 1} IBM Research AI,
\textsuperscript{\rm 2} University of Haifa, Israel,
\textsuperscript{\rm 3} Technion - Israel Institute of Technology\\
\{atereta, boazc, esthergold, amirka, naama.tepper, naamaz\}@il.ibm.com,
gkour@ibm.com,
segevs@technion.ac.il
}

\begin{document}

\maketitle
\begin{abstract}
Based on recent advances in natural language modeling and those in text generation capabilities, we propose a novel data augmentation method for text classification tasks.
We use a powerful pre-trained neural network model to artificially synthesize new labeled data for supervised learning. We mainly focus on cases with scarce labeled data.
Our method, referred to as \term{language-model-based data augmentation (LAMBADA)}, involves fine-tuning a state-of-the-art language generator to a specific task through an initial training phase on the existing (usually small) labeled data.
Using the fine-tuned model and given a class label, new sentences for the class are generated. Our process then filters these new sentences by using a classifier trained on the original data. 
In a series of experiments, we show that LAMBADA improves classifiers’ performance on a variety of datasets.
Moreover, LAMBADA significantly improves upon the state-of-the-art techniques for data augmentation, specifically those applicable to text classification tasks with little data.
% For instance, we achieve absolute improvement of {\color{red} X on Y task and Z on T} task. 
% Furthermore, we show that our new approach is able to promote classifier accuracy and achieve superior results even over naive semi-supervised approaches.
\end{abstract}

\section{Introduction}

\term{Text classification} ~\cite{sebastiani2002machine}, such as  classification of emails into \texttt{spam} and \texttt{not-spam}~\cite{shams2014semi}, is a fundamental research area in machine learning and natural language processing.
It encompasses a variety of other tasks such as intent classification \cite{kumar2019submodular}, sentiment analysis \cite{tang2015document}, topic classification \cite{tong2001support}, and relation classification \cite{giridharastudy}.

Depending upon the problem at hand, getting a good fit for a classifier model may require abundant \term{labeled data}~\cite{shams2014semi}.
However, in many cases, and especially when developing AI systems for specific applications, labeled data is scarce and costly to obtain.

One example of text classification is \term{intent classification} in the growing market of automated chatbot platforms~\cite{collinaszy2017implementation}. 
A developer of an intent classifier for a new chatbot may start with a dataset containing two, three, or five samples per class, and in some cases no data at all.

Data augmentation ~\cite{wong2016understanding} is a common strategy for handling scarce data situations. It works by synthesizing new data from existing training data, with the objective of improving the performance of the downstream model.
This strategy has been a key factor in the performance improvement of various neural network models, mainly in the domains of computer vision and speech recognition.
Specifically, for these domains there exist well-established methods for synthesizing labeled-data to improve classification tasks.
The simpler methods also apply transformations on existing training examples, such as cropping, padding, flipping, and shifting along time and space dimensions, as these transformations are usually class preserving~\cite{krizhevsky2012imagenet,cui2015data,ko2015audio,szegedy2015going}.

However, in the case of \emph{textual} data, such transformations usually invalidate and distort the text, making it grammatically and semantically incorrect. This makes data augmentation more challenging. In fact, textual augmentation could even do more harm than good, since it is \emph{not} an easy task to synthesize good artificial textual data. Thus, data augmentation methods for text usually involve replacing a single word with a synonym, deleting a word, or changing the word order, as suggested by~\cite{wei2019eda}.

Recent advances in text generation models~\cite{radford2018improving,kingma2014auto} facilitate an innovative approach for handling scarce data situations.
Although improving text classification in these situations by using deep learning methods seems like an oxymoron, pre-trained models ~\cite{radford2018improving,peters2018elmo,devlin2019bert} are opening new ways to address this task.  

In this paper, we present a novel method, referred to as \term{language-model-based data augmentation (LAMBADA)}, for synthesizing labeled data to improve text classification tasks. 
LAMBADA is especially useful when only a small amount of labeled data is available, where its results go beyond state-of-the-art performance.
Models trained with LAMBADA exhibit increased performance compared to: \\ 1) The baseline model, trained only on the existing data\\ 2) Models trained on augmented corpora generated by the state-of-the-art techniques in textual data augmentation.

LAMBADA's data augmentation pipeline builds upon a powerful language model: the \term{generative pre-training (GPT)} model~\cite{radford2018improving}.
This neural-network model is pre-trained on huge bodies of text.
As such, it captures the structure of natural language to a great extent, producing deeply coherent sentences and paragraphs. We adapt GPT to our needs by fine-tuning it on the existing, \emph{small}  data. 
We then use the fine-tuned model to synthesize new labeled sentences.
Independently, we train a classifier on the same original small dataset and use it to filter the synthesized data corpus, retaining only data that appears to be qualitative enough.
We then re-train the task classifier on both the existing and the synthesized data. 

We compare LAMBADA to other data augmentation methods and find it statistically better along several datasets and classification algorithms. We mainly focus on small datasets, e.g., containing five examples per class, and show that LAMBADA significantly improves the baseline in such scenarios.

In summary, LAMBADA contributes along three main fronts:
\begin{enumerate}
    \item Statistically improves classifiers’ accuracy. 
    \item Outperforms state-of-the-art data augmentation methods in scarce-data situations.
    \item Suggests a compelling alternative to semi-supervised techniques when unlabeled data does not exist.
\end{enumerate}

The rest of this paper is structured as follows:
In Section~\ref{sec:relatedwork}, we present related work for the state-of-the-art in textual data augmentation techniques and the recent advances in generative pre-trained modeling.
In Section~\ref{sec:problemdifinition}, we define the problem of data augmentation for text classification, and in Section~\ref{sec:ourmethodLAMBADA}, we detail our LAMBADA method solution.
In Section~\ref{sec:results}, we describe the experiments and results  we conducted to analyze LAMBADA performance and to support the paper's main claims. We conclude with a discussion in Section~\ref{sec:discussion}.
 
 % Using classifier results as a means of improving that very same classifier was shown to be effective for \term{semi-supervised learning}~\cite{shams2014semi}, where train data consists of unlabeled data in addition to labeled data. 
\section{Related Work}\label{sec:relatedwork}

Previous textual data augmentation approaches focus on \emph{sample alteration}~\cite{kobayashi2018contextual,wu2019conditional,wei2019eda,mueller2016siamese,jungiewicz2019towards}, in which a single sentence is altered in one way or another, to generate a new sentence while preserving the original class.
One set of these approaches make \emph{local} changes only within a given sentence, primarily by synonym replacement of a word or multiple words.
One of the recent methods in this category is \term{easy data augmentation (EDA)}~\cite{wei2019eda}, which uses simple operations such as synonym replacement and random swap~\cite{miller1995wordnet}.
Another method, \term{conditional BERT contextual augmentation} recently introduced in~\cite{wu2019conditional}, proposes fine-tuned
BERT~\cite{devlin2019bert} for data augmentation by carrying out a masked prediction of words, while conditioning on the class label.
Presumably, methods that make only local changes will produce sentences with a structure similar to the original ones, thus yielding low corpus-level variability.

Other recent possible approaches to textual data augmentation generate whole sentences rather than making a few local changes.
The approaches include using \term{variational autoencoding} (VAE)~\cite{kingma2014auto}, \term{round-trip translation}~\cite{yu2018qanet}, \term{paraphrasing}~\cite{kumar2019submodular}, and methods based on \term{generative adversarial networks}~\cite{tanaka2019data}.
They also include \term{data noising techniques}, such as altering words in the input of self-encoder networks in order to generate a different sentence~\cite{xie2017data,zolna2017fraternal,li2018undeepvo}, or introducing noise on the word-embedding level.
These methods were analyzed in~\cite{marivate2019improving}.
Although a viable option when no access to a formal synonym model exists, they require abundant training data.

Last year, several exciting deep learning methods \cite{vaswani2017attention}
pushed the boundaries of natural language technology. 
They introduced new neural architectures and highly effective transfer learning techniques that dramatically improve natural language processing.
These methods enable the development of new high-performance deep learning models such as \term{ELMO}~\cite{peters2018elmo}, \term{GPT}~\cite{radford2018improving},  \term{BERT}~\cite{devlin2019bert}, and \term{GPT-2}~\cite{radford2019language}.
Common to these models is a \emph{pre-train phase}, in which the models are trained on enormous bodies of publicly available text, and a \emph{fine-tuned phase}, in which they are further trained on task-specific data and loss functions.

When introduced, these models processed natural language better than ever, breaking records in a variety of benchmark tasks related to natural language processing and understanding, as well as tasks involving text generation.
For example, when GPT was first introduced~\cite{radford2018improving}, it improved the state-of-the-art in 12 benchmark tasks, including textual entailment, semantic similarity, sentiment analysis, and commonsense reasoning. These models can produce high-quality sentences even when fine-tuned on small training data. Table~\ref{tab:generated_samples}, shows an example of a few generated sentences based on a small dataset consisting of five sentences per class.

%%%%%%%%%%%%%%%%%%%%%%%%%%
\begin{table}[ht]
\centering
\resizebox{.99\columnwidth}{!}{
\begin{tabular}{||c|l||} 
 \hline
 Class label &  Sentences    \\ 
 \hline\hline
\multirowcell{2}{Flight time}   &  what time is the last flight from san    \\
                           &  francisco to washington dc on continental    \\ 
 \hline

  \multirowcell{2}{Aircraft}   &  show me all the types of aircraft used  \\
                           &  flying from atl to dallas   \\ 
\hline
\multirowcell{2}{City}     &  
                             show me the cities served by canadian \\ & airlines\\ 
 \hline\hline
\end{tabular}
}
\caption{   Examples of generated sentences conditioned on the class label. The generative model was trained on a small dataset consisting of only five sentences per class. }
\label{tab:generated_samples}
\end{table}

%%%%%%%%%%%%%%%%%%%%%%%%%%

These results suggest a counter-intuitive text classification approach: is it possible to fine-tune a pre-trained model and use it to generate new high-quality sentences that will improve the performance of a text classifier?

\section{Problem Definition}\label{sec:problemdifinition}

Text classification is an instance of the \term{supervised learning} problem~\cite{russell2016artificial} over textual data.
In this context, we are given a \term{training dataset} $D_{train} = \{ (x_i, y_i) \}_{i=1}^{n}$ containing $n$ \term{labeled sentences}.
Each $x_i$ is a string of text or, specifically, a sequence of \term{tokens} (roughly, words) $x_i^{1} \ldots x_i^{l_i}$. The label $y_i \in \{1, \ldots, q\}$ indicates the \term{class} of $x_i$ among a set of $q$ classes.
Each $x_i$ is drawn independently from the entire set of strings $X$ (that is, $x_i \in X$), according to an unknown distribution on X, denoted by~$\mathbf{P}_X$.
Moreover, we assume there is an unknown function $f: X \rightarrow \{1, \ldots, q\}$, and that in $D_{train}$, $y_i = f(x_i)$ for all $i=1, \ldots, n$.

The objective of a supervised learning problem is to approximate $f$ on the entire~$X$, given only the dataset~$D_{train}$. In short, we are generalizing from the domain of~$D_{train}$ to the entire~$X$.
Formally, a \term{classification algorithm}~$\mathcal{A}$ receives the training dataset $D_{train}$, and after a training period, it outputs a \term{classifier} function~$h = \mathcal{A}(D_{train})$, where $h: X \rightarrow \{1, \ldots, q\}$ is also known as  a \term{hypothesis}.
To estimate the extent to which $h$ approximates~$f$ on~$X$, it is customary to initially leave out both a training dataset~$D_{train}$ and a \term{test dataset} $D_{test} = \{ (\hat{x}_i, \hat{y}_i) \}_{i=1}^{\hat{n}}$. The test dataset is chosen randomly and has the same structure as $D_{train}$. Both parts are usually drawn from a single, extensive, dataset.

There are different ways of measuring the quality of classifier~$h$ as an approximation to~$f$ using~$D_{test}$.
The most straightforward way measures the \term{accuracy}:
\[
\frac{1}{\hat{n}} \sum_{i=1}^{\hat{n}} \delta(h(\hat{x}_i), \hat{y}_i) ,
\]
where $\delta(\cdot,\cdot)$ is the Kronecker delta (equals~$1$ when both arguments are equal, or~$0$ otherwise), and $\hat{x}_i$, $\hat{y}_i$ are drawn from the test set.
When the test set is large, accuracy approximates the probability of having $h(x)$ equal $f(x)$, namely, $\mathbf{P}_X(h(x) = f(x))$.
We use accuracy as an estimate of classifier performance.

Regardless of how we measure performance, if the train set $D_{train}$ is small, it will dramatically affect the performance of the algorithm~$\mathcal{A}$.
Data augmentation tries to solve this problem by synthesizing additional training pairs that, together with the existing dataset, better reflect the underlying distribution of the data while refraining from introducing too much noise.

Our work does not focus on the classification algorithm per se.
Rather, given a training dataset $D_{train}$ and an algorithm~$\mathcal{A}$, we are interested in a general method for synthesizing an artificial dataset, $D_{synthesized}$.
We aim to apply algorithm~$\mathcal{A}$ on $D_{train} \cup D_{synthesized}$, denoted by $\bar{h} = \mathcal{A}(D_{train} \cup D_{synthesized})$, to yield a relatively good classifier that outperforms the baseline classifier $h$.

In the following section, we describe our method, LAMBADA, and exactly how we obtain  $D_{synthesized}$ from~$D_{train}$ and~$\mathcal{A}$. % by fine-tuning a pre-trained language model and a classifier.
LAMBADA is specifically tailored to the case of small training sets, even miniscule ones, with only a few examples per class.
% Consequently, in Section~\ref{}, GDA is tested on a number of small datasets as well as several classification algorithms.

%Following the notation in \cite{wang2018switchout}, we will use uppercase letters, $X, Y$, to denote random variables and small English letters $x,y$ to denote corresponding actual values.
% The hat sign will denote augmented variables and their values, e.g., $\hat{X}, \hat{Y}, \hat{x}, \hat{y}$, etc.
% Assuming $X$ is the textual instance, i.e., a sequence of words, and $Y \in \mathcal{Y}$ is the corresponding label.
%Given a labeled textual dataset $\mathcal{D}$ containing a small number of samples for each class (in the following we call it the \emph{seed dataset}).

%Using the probabilistic framework we commonly aim at maximizing the following MLE objective:

%\begin{equation}
%     J_{MLE}(\theta)=\mathds{E}_{(x,y)\sim \hat{p}_D}[\log p_\theta (y|x)]
% \label{eq:j_mle}
% \end{equation}
%where $p_\theta(y|x)$ is a parameterized distribution we aim to estimate and $\hat{p}_D$ is the empirical distribution of samples in dataset $D$.

%%%%%%%%%%%%%%%%%%%%%%%%%%%%%%%%%%%%%%%%%%%%%%%%%%%%%%%%%%%%%%%

\section{LAMBADA Method}
\label{sec:ourmethodLAMBADA}

We introduce a novel method for improving the performance of textual classification. Named LAMBADA for its use of Language Model Based Data Augmentation, this method adds more synthesized, weakly-labeled data samples to a given dataset. We define the method in Algorithm~\ref{algo:gda} and elaborate on its steps in the following section. LAMBADA has two key ingredients: 1) model fine-tuning (step 2), which synthesizes labeled data and 2) data filtering (step 4), which retains only high-quality sentences.

\begin{algorithm}
\SetAlgoLined
\SetKwInOut{Input}{Input}
\Input{Training dataset $D_{train}$\\
Classification algorithm $\mathcal{A}$\\
Language model $\mathcal{G}$\\
Number to synthesize per class $N_1, \ldots, N_q$}
Train a baseline classifier $h$ from $D_{train}$ using ~$\mathcal{A}$
\\
Fine-tune $\mathcal{G}$ using~$D_{train}$ to obtain~$\mathcal{G}_{tuned}$
\\
Synthesize a set of labeled sentences $D^*$ using $\mathcal{G}_{tuned}$
\\
Filter $D^*$ using classifier $h$ to obtain $D_{synthesized}$
\\
%Train an extended classifier $\bar{h}$ from $D_{train} \!\cup\! D_{synthesized}$ using algorithm $\mathcal{A}$; i.e., $\bar{h} = \mathcal{A}(D_{train} \cup D_{synthesized})$
%\\
\Return{$D_{synthesized}$} %, $\bar{h}$
\caption{LAMBADA} \label{algo:gda}
\end{algorithm}

\paragraph{Input}

The main input to LAMBADA is a training dataset~$D_{train}$, which we would like to augment with synthesized data. $D_{train}$ contains a set of sentences, each labeled with a class.
To train a classifier, we use a training algorithm~$\mathcal{A}$.
As far as the LAMBADA method is concerned, $\mathcal{A}$ is arbitrary.
However, LAMBADA synthesizes data for the algorithm $\mathcal{A}$, and this is given as a second input to LAMBADA.
This is a distinctive feature of our method. 
We describe both~$D_{train}$ and~$\mathcal{A}$ in Section~\ref{sec:problemdifinition}.

LAMBADA uses a pre-trained \term{language model}~$\mathcal{G}$ to synthesize new data.  
A language model~\cite{bengio2003neural} provides an estimate for the probability that a token (word) will appear, in accordance with a given distribution of text~$\mathbf{P}_{text}$, conditioned on the preceding and/or succeeding tokens. 
More formally, given a token~$w$, and the preceding $k$ tokens (or less) $w^1, \ldots, w^k$, one would like $\mathcal{G}(w | w^1, \ldots, w^k)$ to approximate the conditional probability~$\mathbf{P}_{text}(w | w^1, \ldots, w^k)$ of the appearance of~$w$ in accordance with~$\mathbf{P}_{text}$. $\mathcal{G}$ is usually calculated using a concrete corpus of text~$U$, sampled from distribution~$\mathbf{P}_{text}$.

In contrast to~$\mathcal{A}$, $\mathcal{G}$ is far from being arbitrary. 
We use GPT-2, a recent pre-trained neural-network model (see Section~\ref{sec:relatedwork}), and show that our method outperforms state-of-the-art classifiers in our main use case, where $D_{train}$ is scarce.

GPT-2 is pre-trained on an enormous body of text available on the web. The corpus is organized as a long sequence of tokens, denoted by $U = w^1 \: \cdots \: w^j \: \cdots$.
GPT-2, like GPT, is a right-to-left model based on the \term{transformer} architecture \cite{vaswani2017attention}.
It is pre-trained on~$U$ with loss defined by
\begin{equation} \label{eq:gptloss}
    J_\theta = - \sum_j \log P_\theta(w^j | w^{j-k}, \ldots, w^{j-1}) 
\end{equation}
where $\theta$ is the set of learnable parameters in the neural network of GTP2, and~$P_\theta$ is the trained language model: an estimate of the conditional probability distribution on the set of tokens, as calculated by the network. 
Specifically, we take $\mathcal{G} = P_{\theta^*}$, where~$\theta^*$ indicates the state of the learnable parameters after pre-training.
Nonetheless, from a technical standpoint, the language model and its underlying technology can differ to a great extent, and it is thus presented as a third input to LAMBADA.
As final input in addition to the training set, classification set, and language model LAMBADA is given the number of labeled sentences to synthesize per class~$N_1, \ldots, N_q$.

\paragraph{Step 1: Train baseline classifier}

We train a \term{baseline classifier} $h = \mathcal{A}(D_{train})$ using the existing data $D_{train}$. This classifier will be used for filtering in Step~4.

\paragraph{Step 2: Fine-tune language model}

Independently of Step~1, we fine-tune the language model~$\mathcal{G}$ to the task of synthesizing labeled sentences, to obtain the fine-tuned language model~$\mathcal{G}_{tuned}$.
Here, $\mathcal{G}$ is specifically fine-tuned to the linguistic domain of~$D_{train}$ (that is, the sentences, vocabulary, style, etc.), as well as the particular classes in~$D_{train}$. 
Generally speaking, we would like to use $\mathcal{G}_{tuned}$  to generate a sentence set of any size, and each sentence labeled with a class.

In our case, $\mathcal{G}$ is the neural model of GPT-2.
We fine-tune GPT-2 by training it with the data in~$D_{train} = \{ (x_i, y_i) \}_{i=1}^{n}$.
% However, the training method that we use does \emph{not} need to change.

We concatenate the sentences in~$D_{train}$ in order to form~$U^*$, in the following way:
\begin{equation}\label{eq:ustar}
U^* = y_1 \: \mathtt{SEP} \: x_1 \: \mathtt{EOS} \: y_2 \: \mathtt{SEP} \: x_2 \: \mathtt{EOS} \: y_3 \: \cdots \: y_n \: \mathtt{SEP} \: x_n \: \mathtt{EOS}
\end{equation}
Here, the auxiliary token $\mathtt{SEP}$ separates between a class label and a corresponding sentence, while token $\mathtt{EOS}$ terminates a sentence, and separates it from the label that follows.
We further train the learnable parameters of GPT-2 to predict the next token in the exact same way GPT-2 was pre-trained -- using the loss function in Equation~\ref{eq:gptloss} (with the same training procedure and hyperparameters).
However, we use $U^*$ instead of~$U$, and the learnable parameters are already initialized.
The resulting language model is referred to as~$\mathcal{G}_{tuned}$.

\paragraph{Step 3: Synthesize labeled data}

Given~$\mathcal{G}_{tuned}$, new labeled sentences can be synthesized.
For any class label $y \in \{1, \ldots, q\}$, we can use the adapted language model to predict the continuation of the sequence "$y \: \mathtt{SEP}$" until $\mathtt{EOS}$, which terminates the generated sentence.
This way, for each class, any number of sentences may be synthesized.
For example, this allows us to balance between the classes or otherwise control the ratio of generated sentences per class.
Creating a more balanced training set can improve classification performance, especially in the case of classifiers that are sensitive to unbalanced classes.

In this step, we synthesize a set of labeled sentences, which is denoted by~$D^* = \{(x'_i, y'_i)\}_{i=1}^{N}$. We use a simple and rather crude heuristic, where we generate for each class $y$, 10 times more sentences than we wish to add to the class (i.e., $10 N_y$). Accordingly, the total number of generated sentences is $N = 10 \sum_{y=1}^{q} N_y$. Of course, more sophisticated heuristics can also be examined.
%Similarly, we can generate a set of labeled sentences of any size, by starting from $\mathtt{EOS}$ and letting $\mathcal{G}_{tuned}$ predict tokens indefinitely.
%This sequence of tokens is then parsed according to the structure appearing in Equation~\ref{eq:ustar}, in order to obtain a dataset of labeled sentences, denoted by~$D^* = \{(x'_i, y'_i)\}_{i=1}^{N}$, having the same structure as~$D_{train}$.

GPT-2 generates labeled sentences that are typically both high quality and diverse, facilitating the relative success of our method. This is also where the power of GPT-2 comes into play.
% Note that the distribution of classes may also be taken into account in the generated sequence.

\paragraph{Step 4: Filter synthesized data}

One obstacle in using synthesized text is the noise and error it may introduce.
In the last step, we filter the data in~$D^*$, which was synthesized by~$\mathcal{G}_{tuned}$ in Step~3, leaving only the instances of the highest quality.
We do this using the classifier~$h$ that was trained in Step~1.

For each class~$y$, we take the top~$N_y$ sentences from~$D^*$ that are labeled by~$y$, as follows: 
Given a synthesized sentence $(x, y) \in D^*$, we first verify that~$h(x) = y$, and then use $h$ confidence score (see below) as a rank for $(x,y)$.
That is, we take the top ranked~$N_y$ sentences for class~$y$.
This results in a synthesized dataset $D_{synthesized} \subseteq D^*$, consisting of labeled sentences and with the same structure as~$D_{train}$.
This is the outcome of LAMBADA.

The confidence score given to a data instance by~$h$ can be regarded as the extent the instance is \term{conservative} with respect to~$h$.
In turn, $h$ takes into account both~$D_{train}$ and the algorithm~$\mathcal{A}$ that is to be used with the augmented dataset.
This approach is borrowed from semi-supervised learning~\cite{shams2014semi}, where it is used to classify and filter unlabeled data in a conservative manner.
Note, however, that~$\mathcal{G}_{tuned}$ generates sentences conditioned on a class label.
In our case, this means we have a type of double voting mechanism.

While \emph{not} addressed in this paper, the process described could generally be repeated by applying LAMBADA further on~$D_{train} \cup D_{synthesized}$ to obtain~$D_{synthesized}'$, $D_{synthesized}''$, and so on.

\section{Experimental Results}
\label{sec:results}

We tested our method with three different classifiers (BERT, SVM and LSTM) on three distinct datasets (ATIS, TREC, and WVA) by running multiple experiments in which we varied the amount of training samples per class.
Next, we compared LAMBADA to other data augmentation methods (CVAE, EDA, and CBERT) by using the above-mentioned classifiers and datasets. We statistically validated our results with the McNemar test \cite{mcnemar1947note,dror2018hitchhiker}.

\subsection{Datasets}

Table~\ref{tab:datasets} presents a description of the datasets we used in our experiments.
\begin{table}[ht]
\centering
 \begin{tabular}{||c|c|c|c||} 
 \hline
 Name & Domain & $\#$ Classes & Size \\ [0.5ex] 
 \hline\hline
ATIS & Flight reservations & 17 & 4.2k \\ \hline
TREC & Open-domain questions & 50 &6k \\ \hline
WVA & Telco Customer support & 87& 17k \\ \hline
\end{tabular}
\caption{Datasets.}
\label{tab:datasets}
\end{table}
%%%%%%%%%%%%%%%%%%%%%%%%%%%%%%%%%%%%

\begin{itemize}
    \item \textbf{Airline Travel Information Systems (ATIS)}\footnote{www.kaggle.com/siddhadev/atis-dataset-from-ms-cntk} A dataset providing queries on flight-related information widely used in  language understanding research. ATIS is characterized as an imbalanced dataset, as most of the data belongs to the \textit{flight} class.
    \item 
    \textbf{Text Retrieval Conference (TREC)}\footnote{https://cogcomp.seas.upenn.edu/Data/QA/QC/} A well-known dataset in the information retrieval community for question classification consisting of open-domain, fact-based questions, divided into broad semantic categories.
    \item 
    \textbf{IBM Watson Virtual Assistant (WVA)} A commercial dataset used for intent classification, comprising data for training telco customer support chatbot systems. 
\end{itemize}

We mainly focus on topic classification datasets with the task of classifying a sentence, not an entire document.
Notably, classification of shorter text is considered a more difficult task. We randomly split each dataset into train, validation, and test sets $(80\%, 10\%, 10\%)$.
We then randomly chose from the training set a subset including 5, 10, 20, 50, or 100 samples per class, which we used in each experiment for training. Once determined, we used the same subset throughout all experiments.

\subsection{Classifiers}
We demonstrated that our augmentation approach is independent of the classification algorithm by inspecting three different classifiers, representing three text classification "generations".

\subsubsection{SVM} \term{Support Vector Machine} classifiers were already commonly used before the deep neural network era.
We employ a commercial SVM classifier (IBm Watson Natural Language Classifier) dedicated to natural language processing, which handles both the feature extraction process and the training of the classifier.
While recent models are based on neural networks, in the context of our problem, SVM may have an advantage, since unlike neural-network-based models, it performs well even for relatively small datasets.
    
\subsubsection{LSTM} \term{Long Short Term Memory} represents the type of classifiers that emerged after the advances in training recurrent neural networks, and the introduction of word embeddings \cite{DBLP:journals/corr/abs-1301-3781}, 
LSTMs are commonly used for sequential and textual data classification.
We implemented a sequence-to-vector model based on an LSTM component followed by two fully connected layers and a softmax layer.
For word embedding, we employed GLoVe~\cite{pennington2014glove} of 100 dimensions.
An LSTM classifier usually requires a large amount of data for training.

\subsubsection{BERT} \term{Bidirectional Encoder Representations from Transformers} is a relatively new family of classifiers. Based on the transformer architecture, BERT is pre-trained using two unsupervised tasks: masked language model and next-sentence prediction, on the "BooksCorpus" (800 million words) \cite{zhu2015aligning} and has proven state-of-the-art performance on several text classification tasks. 
Therefore, BERT, like GPT-2, leverages large amounts of data that were used as part of its pre-training phase, in order to perform well, even on relatively small datasets.

\subsection{Generative Models}
\label{subsubsec:generative_models}
We compared LAMBADA's synthetic corpus quality to synthetic corpora generated by various other generative models.
%We compare the ability of other generative models to create qualitative synthetic samples within our framework to the main text generation model used in this paper (GPT-2).
Similar to our selection of classifiers, we selected generators of various types representing different generation approaches.
For a fair comparison, we mainly considered conditional generative models that allow generating samples conditioned on the class label.
This enabled the creation of balanced synthetic corpora, an important feature for some classification models. 
In the following we provide a brief description of these generators.

\subsubsection{EDA} \term{Easy Data Augmentation} \cite{wei2019eda}. This is a recent but simple rule-based data augmentation framework for text. 
 It includes synonym replacement, random insertion, random swap, and random deletion. 
These methods were found beneficial, especially for small training set sizes.

\subsubsection{CVAE}  \term{Conditional Variational Autoencoder} \cite{kingma2014auto}.  This generative model assumes a prior distribution over a latent space and uses deep neural networks to predict its parameters.  
 It is an extension of the Variational Autoencoder model, enabling the conditional generation of an output sentence given a latent vector and the target class. 
We used a standard CVAE model with RNN-based encoder and decoder for generating sentences.

\subsubsection{CBERT} \term{Conditional Bidirectional Encoder Representations from Transformers} \cite{wu2019conditional}. 
As a recent augmentation method for labeled sentences based on BERT, this model operates by randomly replacing words with more varied substitutions predicted by the language model. 
CBERT is pre-trained on a large corpus in an unsupervised setting, allowing it to adapt to specific domains even when fine-tuned through relatively small datasets.

Table \ref{tab:genmodel} describes the attributes of the three generative models mentioned above, including the GPT-2 model.

% benchmark_desc
%%%%%%%%%%%%%%%%%%%%%%%%%%%%%%%%%%%%%%%%%%%%
\begin{table}[ht]
\centering
\begin{tabular}{||c|c|c||}
\hline
Name & Type & External Dataset   \\ \hline \hline
EDA & Rule-Based & Word-Net \\ \hline
CVAE & Autoencoder & -\\ \hline
CBERT & Language Model & Wiki \& Book corpus \\ \hline
GPT-2& Language Model & Web Pages \\ \hline
\end{tabular}
\caption{  Description of the different text generation models.}
\label{tab:genmodel}
\end{table}
%%%%%%%%%%%%%%%%%%%%%%%%%%%%%%%%%%%%%%%%%%%%

\subsection{Results}
We conducted comprehensive experiments, testing LAMBADA's quality from various aspects. We statistically validated all our results with McNemar's test.

\subsubsection{Number of Samples and Classifiers}
We compared the LAMBADA approach with the baseline using three different classifiers over varied numbers of trained samples: 5, 10, 20, 50, and 100 for each class.
We used the ATIS dataset to discover for which sample size our approach is beneficial.

\begin{figure}[ht]
  \centering
  \includegraphics[width=0.95\linewidth]{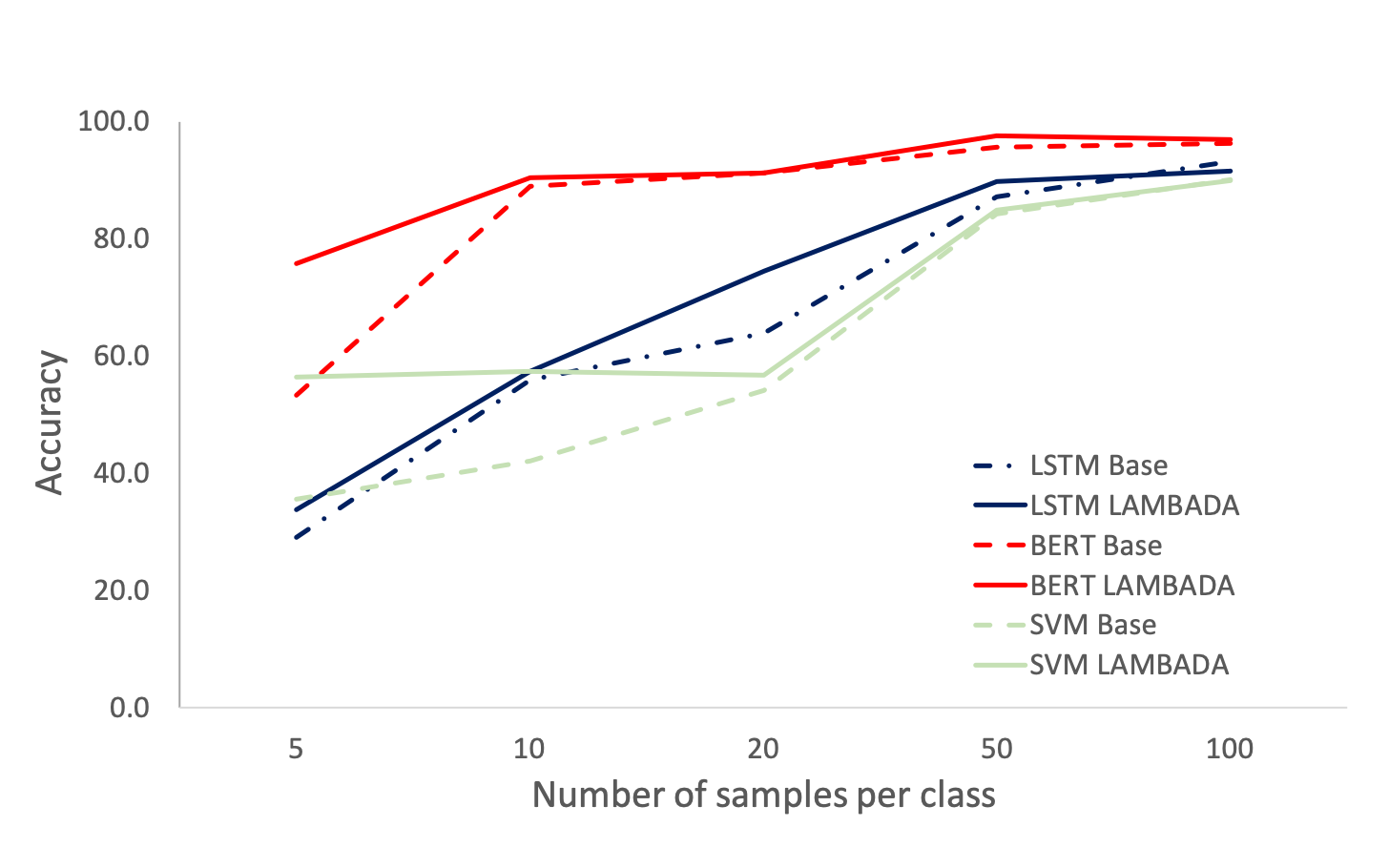}
  \caption{  Accuracy for each sample size over ATIS}
  \label{fig:num_of_samples}
\end{figure}

Figure \ref{fig:num_of_samples} clearly demonstrates the superiority of our LAMBADA approach over the baseline throughout all classifiers and all sample sizes that are smaller than or equal to 50. 
Larger amounts of data do not benefit as much from data augmentation and therefore, in the case of 100 samples for each class, the accuracy of LSTM and SVM does not improve.
%As expected, the gap between the baseline and LAMBADA decreases when enlarging the sample size. 
Figure \ref{fig:num_of_samples} also nicely demonstrates the differences between the classifiers after training them on similar datasets with various sizes.
BERT, which is a pre-trained model, is significantly better than SVM and LSTM throughout all sample sizes. However, the gap between the accuracy of BERT and the other classifiers is more predominant in smaller sample sizes. SVM handles smaller data sizes better than LSTM, as expected. 
Notably, our approach was even able to improve BERT, which is state-of-the-art for text classification and already pre-trained on vast amount of data.

\subsubsection{Datasets} 
We substantiate previous results by comparing the baseline to our LAMBADA approach over three datasets using five samples for each class. Table \ref{tab:classifiers} shows that our approach significantly improves all classifiers over all datasets.

% \input{fig_and_tables/CompareClassifiers5Samples_data.tex}
%%%%%%%%%%%%%%%%%%%%%%%%%%%%%%%%%%%%%%%%%
\begin{table}[ht]
\centering
\begin{tabular}{||c|c|c|c|c||} 
 \hline
 Dataset &   & BERT &  SVM & LSTM     \\ 
 \hline\hline
      \multirowcell{3}{ATIS}  &    {Baseline} & 53.3 & 35.6 & 29.0  \\
                              &  {LAMBADA} & \textbf{75.7} & \textbf{56.5} & \textbf{33.7}  \\ 
                              &  { $\%$ improvement} &   58.5 &   58.7 &    16.2 \\
 \hline\hline
 \multirowcell{3}{TREC} &   {Baseline} & 60.3 & 42.7 & 17.7 \\
                      &  {LAMBADA} & \textbf{64.3} & \textbf{43.9} &  \textbf{25.8} \\
                      &  { $\%$ improvement} &   6.6 &   2.8 &    45.0 \\
 \hline\hline
 \multirowcell{3}{WVA}  &    {Baseline} & 67.2 & 60.2 & 26.0  \\
                      &  {LAMBADA} & \textbf{68.6} & \textbf{62.9} &  \textbf{32.0} \\ 
                      &  { $\%$ improvement} &   2.1 &   4.5 &   23.0 \\
 \hline\hline
\end{tabular}
\caption{Accuracy of LAMBADA vs. baseline over all datasets and classifiers. Significant improvement over all datasets and all classifiers (McNemar, $p-$value$<0.01$).}
\label{tab:classifiers}
\end{table}
%%%%%%%%%%%%%%%%%%%%%%%%%%%%%%%%%%%%%%%%%

Similarly to ATIS dataset,  TREC and WVA datasets also demonstrate the dominance of BERT over SVM and LSTM. LSTM achieves poor results when using a small number of samples, as expected.  
Interestingly, on the ATIS dataset, with BERT and SVM classifiers, the percentage of improvement is far greater than on the other datasets. We believe that this improvement is due to ATIS' imbalanced characteristics and our ability to generate additional data for the under-represented classes.

\subsubsection{Comparison of Generative Models}
We compared our approach to other leading text generator approaches.
Table \ref{tab:banchmark_results} shows that our approach is statistically superior to all other generation algorithms in the ATIS and WVA datasets over all classifiers. In the TREC dataset, the results for BERT are significantly better than all other methods. On the TREC dataset with SVM classifier, our method is on par with EDA.
% Moreover, on the TREC dataset with LSTM classifier, our method is on par with CVAE.

% \input{fig_and_tables/CompareGenerators.tex}
%%%%%%%%%%%%%%%%%%%%%%%%%%%%%%%%%%%%%%%%%%%%%%%%%%
\begin{table}[ht]
\centering
 \begin{tabular}{||c|c|c|c|c||} 
 \hline
 Dataset &  & BERT & SVM & LSTM    \\ 
 \hline\hline
  \multirow{5}*{ATIS} & Baseline & 53.3          & 35.6          & 29.0 \\
                      & EDA      & 62.8          & 35.7          & 27.3    \\ 
                      & CVAE     & 60.6          & 27.6          & 14.9    \\ 
                      & CBERT    & 51.4          & 34.8          & 23.2    \\ 
                      & LAMBADA  & \textbf{75.7}* & \textbf{56.5}* &\textbf{ 33.7}*    \\ 
 \hline\hline
  \multirow{5}*{TREC} & Baseline & 60.3          & 42.7          & 17.7\\
                      & EDA      & 62.6          & \textbf{44.8}*  & 23.1   \\ 
                      & CVAE     & 61.1          & 40.9          & 25.4*    \\ 
                      & CBERT    & 61.4          & 43.8          & 24.2    \\ 
                      & LAMBADA  & \textbf{64.3}* & 43.9*        & \textbf{25.8 }*   \\ 
 \hline\hline
  \multirow{5}*{WVA}  & Baseline & 67.2          & 60.2          & 26.0\\
                      & EDA      & 67.0          & 60.7          & 28.2   \\ 
                      & CVAE     & 65.4          & 54.8          & 22.9    \\ 
                      & CBERT    & 67.4          & 60.7          & 28.4    \\ 
                      & LAMBADA  & \textbf{68.6}* & \textbf{62.9}* & \textbf{32.0}*    \\ 
 \hline\hline
\end{tabular}
\caption{ Accuracy of LAMBADA vs. other generative approaches over all datasets and classifiers. LAMBADA is statistically (* McNemar, $p-$value$<0.01$) superior to all models on each classifier and each dataset (on par to EDA with SVM on TREC).}
\label{tab:banchmark_results}
\end{table}
%%%%%%%%%%%%%%%%%%%%%%%%%%%%%%%%%%%%%%%%%%%%%%

\subsubsection{LAMBADA vs. Unlabeled Data}
Our augmentation framework does not require additional unlabeled data.
As such, it can be applied when unlabeled data is unavailable or costly.
To test the expected LAMBADA performance in such a scenario, we compared it to a semi-supervised approach \cite{ruder2018strong} that uses unlabeled data. Table \ref{tab:unlabeled} compares between different experimental settings on ATIS using three classifiers and five samples per class.

% \input{fig_and_tables/Unlabeled.tex}
%%%%%%%%%%%%%%%%%%%%%%%%%%%%%%%%%%%%%%%%
\begin{table}[ht]
\centering
\begin{tabular}{||c||c|c|c|c||}
\hline
 Classifier & Base. & Unlab. & Unlab. & LAMBADA  \\ 
           &           & Data      &    GPT     &        \\ \hline \hline
BERT        & 53.3      & 54.5              & 73.2      & \textbf{75.7*} \\ \hline
SVM         & 35.6      & 23.5              & 47.2      & \textbf{56.5*} \\ \hline
LSTM        & 29.0      & \textbf{40.1*}     & 23.2      & 33.7          \\ \hline

\end{tabular}
\caption{ Accuracy of LAMBADA with or without label vs. unlabeled data for ATIS dataset with 5 samples per class. Significant improvement for BERT and SVM classifiers (*McNemar, $p-$value$<0.01$).}
\label{tab:unlabeled}
\end{table}
%%%%%%%%%%%%%%%%%%%%%%%%%%%%%%%%%%%%%%%%

To create an \textit{unlabeled} dataset, we randomly selected samples from the original dataset while ignoring their labels.
Next, following a simple \emph{weak labeling} approach, we classified the samples with one of the classifiers after training it on the labeled dataset.
We compared LAMBADA's classification results with the results we obtained from this classifier.
These results appear in the \textit{LAMBADA} and \textit{Unlabeled data} columns of Table \ref{tab:unlabeled}.
Surprisingly, for most classifiers, LAMBADA achieves better accuracy compared to a simple weak labeling approach.
Clearly, the generated dataset contributes more to improving the accuracy of the classifier than the unlabeled samples taken from the original dataset.

We may attribute this increased performance to two main factors:
\begin{enumerate}
    \item LAMBADA uses its "generated" labels, which significantly improve performance.
    \item
    LAMBADA allows us to control the number of samples per class by investing more effort in generating samples for classes that are under-represented in the original dataset.
\end{enumerate}

We further assessed the importance of the "generated" labels by removing them from LAMBADA’s synthesized dataset. We provide the results for this experiment under the \textit{GPT Unlabeled} column in Table \ref{tab:unlabeled}. In future work, we plan to use various data balancing approaches on the unlabeled dataset to assess the importance of the second factor above.

\section{Discussion and Future Work}
\label{sec:discussion}
We introduce LAMBADA for improving classifiers' performance. It involves fine-tuning a language model, generating new labeled-condition sentences and a filtering phase. We showed that our method statically improves classifiers' performance on small data sets. In addition, we showed that LAMBADA beats the state-of-the-art techniques in data augmentation.

\subsubsection{Generative vs. Discriminative} Generally speaking, training a generative model requires more data than training a discriminative model \cite{ng2002discriminative}.
This is attributed mainly to the fact that discriminative models aim at estimating the class boundaries, while generative models approximate the probability distribution of the samples in each class.
Therefore, prima facie, it is counter-intuitive to employ a generative model to improve discriminative classifier accuracy. All the more so, when both models are trained on the same dataset.
However, unlike discriminative models, generative models may exploit unsupervised data to compensate for the inherent higher sample complexity.
Consequently, and given the available abundant amount of unlabeled textual data, language models, pre-trained on huge corpora, have recently become state-of-the-art.
Fine-tuning these generative models requires an extremely small amount of labeled data, as we show in this work, and sampling from them is straightforward.

\subsubsection{Filtering Approach}  LAMBADA synthesizes data in two steps. It first generates a large number of sentences per class and then filters them by multiple conditions. In this work, we employ a simple filtering heuristic, inspired by the semi-supervised paradigm that takes into account: 1) the class label of the generated sentence 2) the class label as given by the filtering classifier, together with its confidence score and 3) the number of sentences per class.  We plan to further investigate other filtering heuristics and approaches in future work.

\subsubsection{Weak Labeling and Self-Supervision} LAMBADA synthesizes corpora of weakly labeled data by conditionally generating sentences on a given class' label. Thus, one may incorporate a LAMBADA synthesized corpus within any weak labeling or semi-supervised framework such as one of these suggested by \cite{ruder2018strong}. Moreover, one may use a synthesized corpus in situations where unlabeled data is not available and still expect comparable results.  
\subsubsection{Zero-shot Learning} Most textual datasets contain class names with semantic meaning. 
LAMBADA, an approach based on a language model, utilizes this class label meaning in its generation process. 
Consequently, it enables synthesizing samples for any meaningful, domain-related, class name. It thus potentially allows the generation of samples for unseen classes, a method also known as zero-shot learning \cite{socher2013zero}.
We plan to investigate this idea in future research.

 \subsubsection{Iterative Training Process} While a single step of the augmentation process may sufficiently improve the classifier, as shown in this paper, there is no real impediment to repeat the process by running several iterations of Algorithm~\ref{algo:gda}.
One of the possible hazards that the repetition of this process may cause is \emph{data drifting}, in which biased synthesized samples gain domination over the training dataset.

\fontsize {9.0pt}{10.0pt} \selectfont
\bibliographystyle{aaai}
\bibliography{AAAI-AnabyA.4027.bib}

\begin{thebibliography}{}

\bibitem[\protect\citeauthoryear{Bengio \bgroup et al\mbox.\egroup
  }{2003}]{bengio2003neural}
Bengio, Y.; Ducharme, R.; Vincent, P.; and Jauvin, C.
\newblock 2003.
\newblock A neural probabilistic language model.
\newblock {\em Journal of machine learning research} 3(Feb):1137--1155.

\bibitem[\protect\citeauthoryear{Collinaszy, Bundzel, and
  Zolotova}{2017}]{collinaszy2017implementation}
Collinaszy, J.; Bundzel, M.; and Zolotova, I.
\newblock 2017.
\newblock Implementation of intelligent software using ibm watson and bluemix.
\newblock {\em Acta Electrotechnica et Informatica} 17(1):58--63.

\bibitem[\protect\citeauthoryear{Cui, Goel, and Kingsbury}{2015}]{cui2015data}
Cui, X.; Goel, V.; and Kingsbury, B.
\newblock 2015.
\newblock Data augmentation for deep neural network acoustic modeling.
\newblock {\em IEEE/ACM Transactions on Audio, Speech and Language Processing
  (TASLP)} 23(9):1469--1477.

\bibitem[\protect\citeauthoryear{Devlin \bgroup et al\mbox.\egroup
  }{2019}]{devlin2019bert}
Devlin, J.; Chang, M.-W.; Lee, K.; and Toutanova, K.
\newblock 2019.
\newblock Bert: Pre-training of deep bidirectional transformers for language
  understanding.
\newblock In {\em Proceedings of the 2019 Conference of the North American
  Chapter of the Association for Computational Linguistics: Human Language
  Technologies, Volume 1 (Long and Short Papers)},  4171--4186.

\bibitem[\protect\citeauthoryear{Dror \bgroup et al\mbox.\egroup
  }{2018}]{dror2018hitchhiker}
Dror, R.; Baumer, G.; Shlomov, S.; and Reichart, R.
\newblock 2018.
\newblock The hitchhiker’s guide to testing statistical significance in
  natural language processing.
\newblock In {\em Proceedings of the 56th Annual Meeting of the Association for
  Computational Linguistics (Volume 1: Long Papers)},  1383--1392.

\bibitem[\protect\citeauthoryear{Giridhara, Mishra, and
  Modam}{2019}]{giridharastudy}
Giridhara, P. K.~B.; Mishra, C.; and Modam, R.~K.
\newblock 2019.
\newblock A study of various text augmentation techniques for relation
  classification in free text.
\newblock In {\em Proceedings of the 8th International Conference on Pattern
  Recognition Applications and Methods}.

\bibitem[\protect\citeauthoryear{Jungiewicz and
  Smywinski-Pohl}{2019}]{jungiewicz2019towards}
Jungiewicz, M., and Smywinski-Pohl, A.
\newblock 2019.
\newblock Towards textual data augmentation for neural networks: synonyms and
  maximum loss.
\newblock {\em Computer Science} 20(1).

\bibitem[\protect\citeauthoryear{Kingma and Welling}{2014}]{kingma2014auto}
Kingma, D.~P., and Welling, M.
\newblock 2014.
\newblock Auto-encoding variational bayes.
\newblock {\em stat} 1050:10.

\bibitem[\protect\citeauthoryear{Ko \bgroup et al\mbox.\egroup
  }{2015}]{ko2015audio}
Ko, T.; Peddinti, V.; Povey, D.; and Khudanpur, S.
\newblock 2015.
\newblock Audio augmentation for speech recognition.
\newblock In {\em Sixteenth Annual Conference of the International Speech
  Communication Association}.

\bibitem[\protect\citeauthoryear{Kobayashi}{2018}]{kobayashi2018contextual}
Kobayashi, S.
\newblock 2018.
\newblock Contextual augmentation: Data augmentation by words with paradigmatic
  relations.
\newblock In {\em Proceedings of the 2018 Conference of the North American
  Chapter of the Association for Computational Linguistics: Human Language
  Technologies, Volume 2 (Short Papers)},  452--457.

\bibitem[\protect\citeauthoryear{Krizhevsky, Sutskever, and
  Hinton}{2012}]{krizhevsky2012imagenet}
Krizhevsky, A.; Sutskever, I.; and Hinton, G.~E.
\newblock 2012.
\newblock Imagenet classification with deep convolutional neural networks.
\newblock In {\em Advances in neural information processing systems},
  1097--1105.

\bibitem[\protect\citeauthoryear{Kumar \bgroup et al\mbox.\egroup
  }{2019}]{kumar2019submodular}
Kumar, A.; Bhattamishra, S.; Bhandari, M.; and Talukdar, P.
\newblock 2019.
\newblock Submodular optimization-based diverse paraphrasing and its
  effectiveness in data augmentation.
\newblock In {\em Proceedings of the 2019 Conference of the North American
  Chapter of the Association for Computational Linguistics: Human Language
  Technologies, Volume 1 (Long and Short Papers)},  3609--3619.

\bibitem[\protect\citeauthoryear{Li \bgroup et al\mbox.\egroup
  }{2018}]{li2018undeepvo}
Li, R.; Wang, S.; Long, Z.; and Gu, D.
\newblock 2018.
\newblock Undeepvo: Monocular visual odometry through unsupervised deep
  learning.
\newblock In {\em 2018 IEEE International Conference on Robotics and Automation
  (ICRA)},  7286--7291.
\newblock IEEE.

\bibitem[\protect\citeauthoryear{Marivate and
  Sefara}{2019}]{marivate2019improving}
Marivate, V., and Sefara, T.
\newblock 2019.
\newblock Improving short text classification through global augmentation
  methods.
\newblock {\em arXiv preprint arXiv:1907.03752}.

\bibitem[\protect\citeauthoryear{McNemar}{1947}]{mcnemar1947note}
McNemar, Q.
\newblock 1947.
\newblock Note on the sampling error of the difference between correlated
  proportions or percentages.
\newblock {\em Psychometrika} 12(2):153--157.

\bibitem[\protect\citeauthoryear{Mikolov \bgroup et al\mbox.\egroup
  }{2013}]{DBLP:journals/corr/abs-1301-3781}
Mikolov, T.; Chen, K.; Corrado, G.; and Dean, J.
\newblock 2013.
\newblock Efficient estimation of word representations in vector space.
\newblock {\em CoRR} abs/1301.3:1--12.

\bibitem[\protect\citeauthoryear{Miller}{1995}]{miller1995wordnet}
Miller, G.~A.
\newblock 1995.
\newblock Wordnet: a lexical database for english.
\newblock {\em Communications of the ACM} 38(11):39--41.

\bibitem[\protect\citeauthoryear{Mueller and
  Thyagarajan}{2016}]{mueller2016siamese}
Mueller, J., and Thyagarajan, A.
\newblock 2016.
\newblock Siamese recurrent architectures for learning sentence similarity.
\newblock In {\em Thirtieth AAAI Conference on Artificial Intelligence}.

\bibitem[\protect\citeauthoryear{Ng and Jordan}{2002}]{ng2002discriminative}
Ng, A.~Y., and Jordan, M.~I.
\newblock 2002.
\newblock On discriminative vs. generative classifiers: A comparison of
  logistic regression and naive bayes.
\newblock In {\em Advances in neural information processing systems},
  841--848.

\bibitem[\protect\citeauthoryear{Pennington, Socher, and
  Manning}{2014}]{pennington2014glove}
Pennington, J.; Socher, R.; and Manning, C.
\newblock 2014.
\newblock Glove: Global vectors for word representation.
\newblock In {\em Proceedings of the 2014 conference on empirical methods in
  natural language processing (EMNLP)},  1532--1543.

\bibitem[\protect\citeauthoryear{Peters \bgroup et al\mbox.\egroup
  }{2018}]{peters2018elmo}
Peters, M.~E.; Neumann, M.; Iyyer, M.; Gardner, M.; Clark, C.; Lee, K.; and
  Zettlemoyer, L.
\newblock 2018.
\newblock Deep contextualized word representations.
\newblock In {\em Proc. of NAACL}.

\bibitem[\protect\citeauthoryear{Radford \bgroup et al\mbox.\egroup
  }{2018}]{radford2018improving}
Radford, A.; Narasimhan, K.; Salimans, T.; and Sutskever, I.
\newblock 2018.
\newblock Improving language understanding by generative pre-training.
\newblock {\em URL https://s3-us-west-2. amazonaws.
  com/openai-assets/research-covers/languageunsupervised/language understanding
  paper. pdf}.

\bibitem[\protect\citeauthoryear{Radford \bgroup et al\mbox.\egroup
  }{2019}]{radford2019language}
Radford, A.; Wu, J.; Child, R.; Luan, D.; Amodei, D.; and Sutskever, I.
\newblock 2019.
\newblock Language models are unsupervised multitask learners.
\newblock {\em OpenAI Blog} 1:8.

\bibitem[\protect\citeauthoryear{Ruder and Plank}{2018}]{ruder2018strong}
Ruder, S., and Plank, B.
\newblock 2018.
\newblock Strong baselines for neural semi-supervised learning under domain
  shift.
\newblock In {\em The 56th Annual Meeting of the Association for Computational
  LinguisticsMeeting of the Association for Computational Linguistics}.
\newblock Association for Computational Linguistics.

\bibitem[\protect\citeauthoryear{Russell and
  Norvig}{2016}]{russell2016artificial}
Russell, S.~J., and Norvig, P.
\newblock 2016.
\newblock {\em Artificial intelligence: a modern approach}.
\newblock Malaysia; Pearson Education Limited,.

\bibitem[\protect\citeauthoryear{Sebastiani}{2002}]{sebastiani2002machine}
Sebastiani, F.
\newblock 2002.
\newblock Machine learning in automated text categorization.
\newblock {\em ACM computing surveys (CSUR)} 34(1):1--47.

\bibitem[\protect\citeauthoryear{Shams}{2014}]{shams2014semi}
Shams, R.
\newblock 2014.
\newblock Semi-supervised classification for natural language processing.
\newblock {\em arXiv preprint arXiv:1409.7612}.

\bibitem[\protect\citeauthoryear{Socher \bgroup et al\mbox.\egroup
  }{2013}]{socher2013zero}
Socher, R.; Ganjoo, M.; Manning, C.~D.; and Ng, A.
\newblock 2013.
\newblock Zero-shot learning through cross-modal transfer.
\newblock In {\em Advances in neural information processing systems},
  935--943.

\bibitem[\protect\citeauthoryear{Szegedy \bgroup et al\mbox.\egroup
  }{2015}]{szegedy2015going}
Szegedy, C.; Liu, W.; Jia, Y.; Sermanet, P.; Reed, S.; Anguelov, D.; Erhan, D.;
  Vanhoucke, V.; and Rabinovich, A.
\newblock 2015.
\newblock Going deeper with convolutions.
\newblock In {\em Proceedings of the IEEE conference on computer vision and
  pattern recognition},  1--9.

\bibitem[\protect\citeauthoryear{Tanaka and Aranha}{2019}]{tanaka2019data}
Tanaka, F. H. K. d.~S., and Aranha, C.
\newblock 2019.
\newblock Data augmentation using gans.
\newblock {\em arXiv preprint arXiv:1904.09135}.

\bibitem[\protect\citeauthoryear{Tang, Qin, and Liu}{2015}]{tang2015document}
Tang, D.; Qin, B.; and Liu, T.
\newblock 2015.
\newblock Document modeling with gated recurrent neural network for sentiment
  classification.
\newblock In {\em Proceedings of the 2015 conference on empirical methods in
  natural language processing},  1422--1432.

\bibitem[\protect\citeauthoryear{Tong and Koller}{2001}]{tong2001support}
Tong, S., and Koller, D.
\newblock 2001.
\newblock Support vector machine active learning with applications to text
  classification.
\newblock {\em Journal of machine learning research} 2(Nov):45--66.

\bibitem[\protect\citeauthoryear{Vaswani \bgroup et al\mbox.\egroup
  }{2017}]{vaswani2017attention}
Vaswani, A.; Shazeer, N.; Parmar, N.; Uszkoreit, J.; Jones, L.; Gomez, A.~N.;
  Kaiser, {\L}.; and Polosukhin, I.
\newblock 2017.
\newblock Attention is all you need.
\newblock In {\em Advances in neural information processing systems},
  5998--6008.

\bibitem[\protect\citeauthoryear{Wei and Zou}{2019}]{wei2019eda}
Wei, J.~W., and Zou, K.
\newblock 2019.
\newblock Eda: Easy data augmentation techniques for boosting performance on
  text classification tasks.
\newblock In {\em Seventh International Conference on Learning
  Representations}, ICLR 2019 Workshop LLD.

\bibitem[\protect\citeauthoryear{Wong \bgroup et al\mbox.\egroup
  }{2016}]{wong2016understanding}
Wong, S.~C.; Gatt, A.; Stamatescu, V.; and McDonnell, M.~D.
\newblock 2016.
\newblock Understanding data augmentation for classification: when to warp?
\newblock In {\em 2016 international conference on digital image computing:
  techniques and applications (DICTA)},  1--6.
\newblock IEEE.

\bibitem[\protect\citeauthoryear{Wu \bgroup et al\mbox.\egroup
  }{2019}]{wu2019conditional}
Wu, X.; Lv, S.; Zang, L.; Han, J.; and Hu, S.
\newblock 2019.
\newblock Conditional bert contextual augmentation.
\newblock In {\em International Conference on Computational Science},  84--95.
\newblock Springer.

\bibitem[\protect\citeauthoryear{Xie \bgroup et al\mbox.\egroup
  }{2017}]{xie2017data}
Xie, Z.; Wang, S.~I.; Li, J.; L{\'e}vy, D.; Nie, A.; Jurafsky, D.; and Ng,
  A.~Y.
\newblock 2017.
\newblock Data noising as smoothing in neural network language models.
\newblock In {\em Proceedings of the Fifth International Conference on Learning
  Representations}.

\bibitem[\protect\citeauthoryear{Yu \bgroup et al\mbox.\egroup
  }{2018}]{yu2018qanet}
Yu, A.~W.; Dohan, D.; Luong, M.-T.; Zhao, R.; Chen, K.; Norouzi, M.; and Le,
  Q.~V.
\newblock 2018.
\newblock Qanet: Combining local convolution with global self-attention for
  reading comprehension.
\newblock In {\em 6th International Conference on Learning Representations
  (ICLR 2018)}.

\bibitem[\protect\citeauthoryear{Zhu \bgroup et al\mbox.\egroup
  }{2015}]{zhu2015aligning}
Zhu, Y.; Kiros, R.; Zemel, R.; Salakhutdinov, R.; Urtasun, R.; Torralba, A.;
  and Fidler, S.
\newblock 2015.
\newblock Aligning books and movies: Towards story-like visual explanations by
  watching movies and reading books.
\newblock In {\em Proceedings of the IEEE international conference on computer
  vision},  19--27.

\bibitem[\protect\citeauthoryear{Zolna \bgroup et al\mbox.\egroup
  }{2017}]{zolna2017fraternal}
Zolna, K.; Arpit, D.; Suhubdy, D.; and Bengio, Y.
\newblock 2017.
\newblock Fraternal dropout.
\newblock In {\em Poster presented at the Sixth International Conference on
  Learning Representations}.

\end{thebibliography}

\end{document}